\documentclass[runningheads]{llncs}

\usepackage{times}
\usepackage{epsfig}
\usepackage{graphicx}
\usepackage{amsmath}
\usepackage{amssymb}
\usepackage{amsfonts}
\usepackage{stmaryrd}
\usepackage{algorithm}
\usepackage{algpseudocode}
\usepackage{xcolor}
\usepackage[skip=0cm,list=true,labelfont=it]{subcaption}
\usepackage{wrapfig}

\usepackage{url}
\usepackage{hyperref}

\begin{document}

\title{UCSL : A Machine Learning Expectation-Maximization framework for Unsupervised Clustering driven by Supervised Learning}
%
\titlerunning{UCSL}

\author{
Robin Louiset\inst{1,2} \and
Pietro Gori\inst{2} \and
Benoit Dufumier\inst{1,2} \and
Josselin Houenou\inst{1} \and
Antoine Grigis\inst{1} \and
Edouard Duchesnay\inst{1}
}
%
\authorrunning{R. Louiset et al.}
%
\institute{
Université Paris-Saclay, CEA, Neurospin, 91191, Gif-sur-Yvette, France \and
LTCI, Télécom Paris, Institut Polytechnique de Paris, France
}
%
\maketitle              
\begin{abstract}

Subtype Discovery consists in finding interpretable and consistent sub-parts of a dataset, which are also relevant to a certain supervised task. From a mathematical point of view, this can be defined as a clustering task driven by supervised learning in order to uncover subgroups in line with the supervised prediction.
In this paper, we propose a general Expectation-Maximization ensemble framework entitled UCSL (Unsupervised Clustering driven by Supervised Learning). Our method is generic, it can integrate any clustering method and can be driven by both binary classification and regression. We propose to construct a non-linear model by merging multiple linear estimators, one per cluster. Each hyperplane is estimated so that it correctly discriminates - or predict - only one cluster. We use SVC or Logistic Regression for classification and SVR for regression.
Furthermore, to perform cluster analysis within a more suitable space, we also propose a dimension-reduction algorithm that projects the data onto an orthonormal space relevant to the supervised task.
We analyze the robustness and generalization capability of our algorithm using synthetic and experimental datasets. In particular, we validate its ability to identify suitable consistent sub-types by conducting a psychiatric-diseases cluster analysis with known ground-truth labels. The gain of the proposed method over previous state-of-the-art techniques is about +1.9 points in terms of balanced accuracy.
Finally, we make codes and examples available in a scikit-learn-compatible Python package.
\url{https://github.com/neurospin-projects/2021_rlouiset_ucsl/}
\end{abstract}
%
%
%

\section{Introduction}

\begin{wrapfigure}[10]{i}{0.4\textwidth}
  \vspace{-5mm}
    \includegraphics[width=0.38\textwidth]{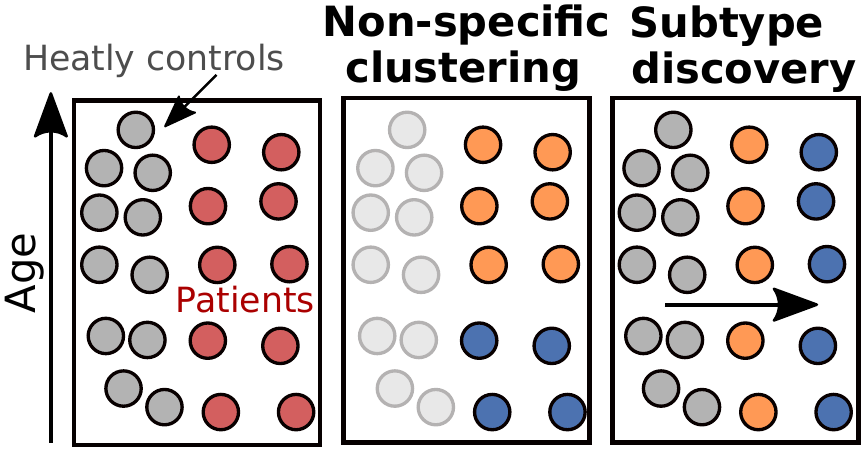}
  \caption{Subtype discovery in clinical research.}
  \label{fig:ucsl_principles}
\end{wrapfigure}

Subtype discovery is the task of finding consistent subgroups within a population or a class of objects which are also relevant to a certain supervised upstream task. This means that the definition of homogeneity of subtypes should not be fully unsupervised, as in standard clustering, but it should also be driven by a supervised task. For instance, when identifying flowers, one may want to find different varieties or subtypes within each species. Standard clustering algorithms are driven by features that explain most of the general variability, such as the height or the thickness. Subtype identification aims at discovering subgroups describing the specific heterogeneity within each flower species and not the general variability of flowers.
To disentangle these sources of variability, a supervised task can identify a more relevant feature space to drive the intra-species clustering problem.
Depending on the domain, finding relevant subgroups may turn out to be a relatively hard task. Indeed, most of the time, boundaries between different patterns are fuzzy and may covariate with other factors. Hence, ensuring that resulting predictions are not collapsed clusters or biased by an irrelevant confound factor is a key step in the development of such analysis.
For example, in clinical research, it is essential to identify subtypes of patients with a given disorder (red dots in Fig.~\ref{fig:ucsl_principles}). The problem is that the general variability (that stems from age or sex) is observed in both healthy controls (grey dots in Fig.~\ref{fig:ucsl_principles}) and disease patients, therefore it will probably drive the clustering of patients toward a non-specific solution (second plot in Fig.~\ref{fig:ucsl_principles}). Adding a supervised task (healthy controls vs patients) can be used to find direction(s) (horizontal arrow Fig.~\ref{fig:ucsl_principles}) that discards non-specific variability to emphasize more disease-related differences (subtype discovery in Fig.~\ref{fig:ucsl_principles}).
This is a fundamental difference between unsupervised clustering analysis and subtype identification.

Subgroups identification is highly relevant in various fields such as in clinical research where disease subtypes discovery can lead to better personalized drug-treatment and prognosis \cite{wu_cancer_2013} or to better anticipate at-risk profiles \cite{wang_unsupervised_2020}. Particularly, given the extreme variability of cancer, identifying subtypes enable to develop precision medicine \cite{carey_race_2006,marusyk_tumor_2010,menyhart_multi-omics_2021,wu_cancer_2013,planey_coincide_2016,wu_cancer_2013}. In psychiatry and neurology, different behaviour, anatomical and physiological patterns point out variants of mental disorders \cite{marquand_beyond_2016} such as for bipolar disorder \cite{yang_probing_2021}, schizophrenia \cite{honnorat_neuroanatomical_2019,chand_two_2020}, autism, \cite{zabihi_dissecting_2019,tager-flusberg_identifying_2003}, attention-deficit hyperactivity disorder \cite{wahlstedt_heterogeneity_2009}, Alzheimer's disease \cite{ferreira_distinct_2017,HYDRA,yang_smile-gans_2020,wen_magic_2020} or Parkinson's disease \cite{erro_heterogeneity_2013}. In bio-informatics, DNA subfolds analysis is a key field for the understanding of gene functions and regulations, cellular processes and cells subtyping \cite{sonpatki_recursive_2020}. In the field of data mining, crawling different consistent subgroups of written data enables enhanced applications \cite{rawat_hybrid_2019}.

\section{Related works}

Early works \cite{carey_race_2006,erro_heterogeneity_2013} proposed traditional clustering methods to find relevant subgroups for clinical research in cancer and neurology. However, they were very sensitive to high-dimensional data and noise, making them hardly reproducible \cite{oyelade_clustering_2016,planey_coincide_2016}. 

To overcome these limits, \cite{sonpatki_recursive_2020} and \cite{planey_coincide_2016} evaluated custom consensus methods to fuse multiple clustering estimates in order to obtain more robust and reproducible results. Additionally, \cite{sonpatki_recursive_2020} also proposed to select the most important features in order to overcome the curse of dimensionality. 
Even if all these methods provide relevant strategies to identify stable clusters in high-dimensional space, they do not allow the identification of disease-specific subtypes when the dominant variability in patients corresponds to the variability in the general population. To select disease-specific variability, recent contributions propose hybrid approaches integrating a supervised task (patient vs. controls) to the clustering problem.
In \cite{schulz_inferring_2020}, authors propose a hybrid method for disease-subtyping in precision medicine. Their implementation consists of training a Random Forest supervised classifier (healthy vs. diseased) and then apply SHAP algorithm \cite{lundberg_unified_2017,lundberg_consistent_2017} to get explanation values from Random Forest classifiers. This yields promising results even though it is computationally expensive, especially when the dataset size increases.

Differently, a wide range of Deep Learning methods propose to learn better representations via deep encoders and adapt clustering method on compressed latent space or directly within the minimizing loss. In this case, encoders have to be trained with at least one non-clustering loss, to enhance the representations \cite{saito_neural_2017} and avoid collapsing clusters \cite{yang_towards_2017}. \cite{caron_deep_2018} proposes a Deep Clustering framework that alternates between latent clusters estimation and likelihood maximization through pseudo-label classification. Yet, its training remains unstable and designed for large-scale dataset only. Prototypical Contrastive Learning \cite{li_prototypical_2021}, SeLA \cite{asano_self-labelling_2020}, SwAV \cite{caron_unsupervised_2020} propose contrastive learning frameworks that alternatively maximize 1- the mutual information between the input samples and their latent representations and 2- the clustering estimation. These works have proven to be very efficient and stable on large-scale datasets. They compress inputs into denser and richer representations, and successfully get rid of unnecessary noisy dimensions. Nevertheless, they still do not propose a representation aligned with the supervised task at-hand. To ensure that resulting clusters identify relevant subgroups for the supervised task,
one could first train for the supervised task and then run clustering on the latent space. This would emphasize important features for the supervised task but it may also regress out intra-class specific heterogeneity, hence the need of an iterative process where clustering and classification tasks influence each other. 

\begin{figure}[!tbp]
\centering
\includegraphics[width=\textwidth]{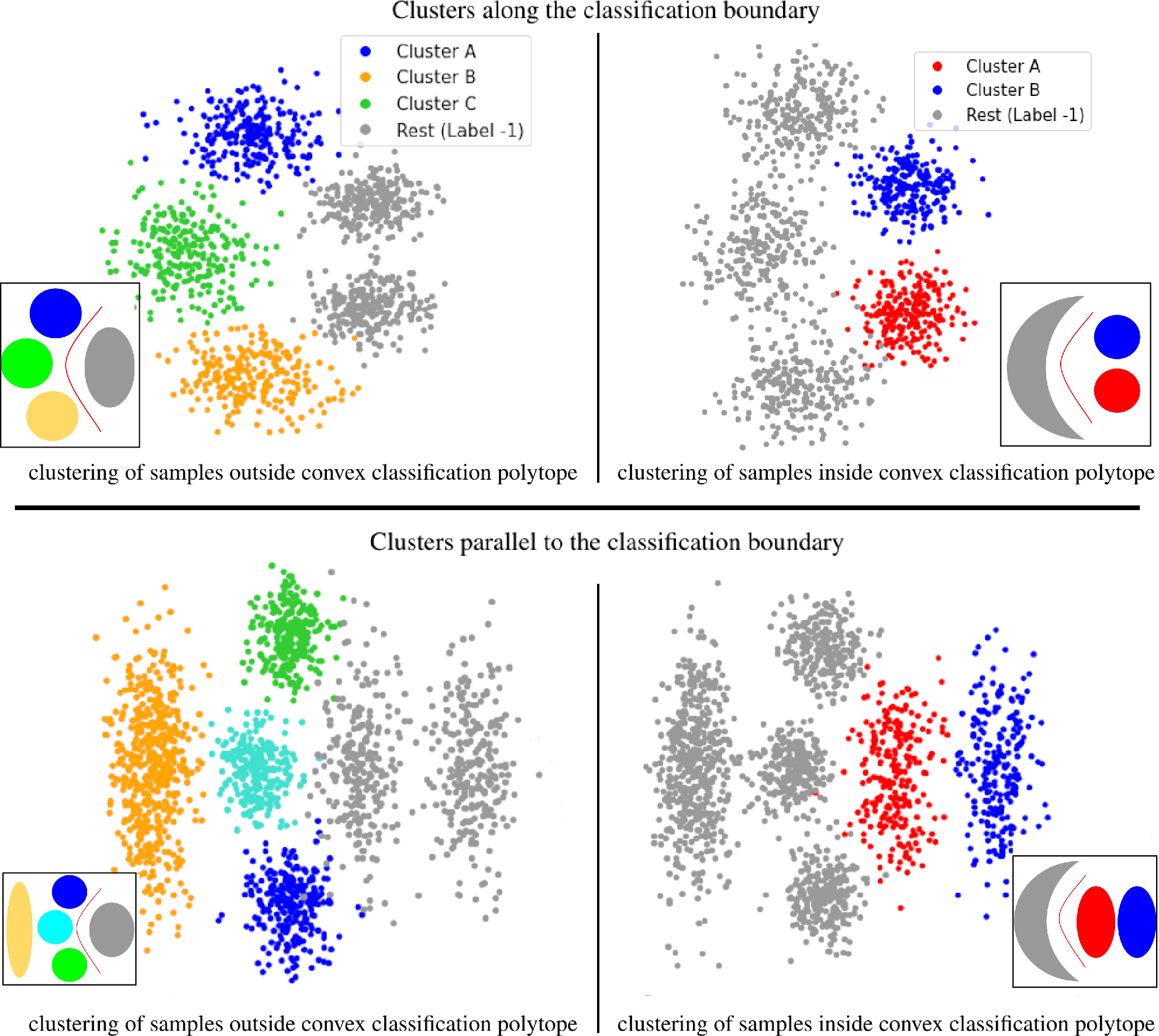}
\centering
\caption{Toy Datasets - Different configurations we want to address. Grey points represent negative samples. The upstream task is to classify negative (grey) samples from all positive (colored) samples while the final goal is to cluster positive samples. The upper plots show 3 and 2 clusters respectively along the classification boundary. The lower plot show 4 and 2 clusters respectively parallel (and also along on the left) to the classification boundary. Furthermore, plots on the left and right show clusters outside and inside the convex classification polytope respectively.}
\label{fig:toy_examples_configuration_table}
\end{figure}

CHIMERA \cite{honnorat_neuroanatomical_2019}, proposes an Alzheimer's subtype discovery algorithm driven by supervised classification between healthy and pathological samples. It assumes that the pathological heterogeneity can be modeled as a set of linear transformations from the reference set of healthy subjects to the patient distribution, where each transformation corresponds to one pathological subtype. This is a strong a priori that limits its application to (healthy reference)/(pathological case) only.
\cite{HYDRA,wen_magic_2020} propose an alternate algorithm between supervised learning and unsupervised cluster analysis where each step influences the other until it reaches a stable configuration. The algorithm simultaneously solves binary classification and intra-class clustering in a hybrid fashion thanks to a maximum margin framework. The method discriminates healthy controls from pathological patients by optimizing the best convex polytope that is formed by combining several linear hyperplanes. The clustering ability is drawn by assigning patients to their best discriminating hyperplane. Each cluster corresponds to one face of the piece-wise linear polytope and heterogeneity is implicitly captured by harnessing the classification boundary non-linearity.
The efficiency of this method heavily relies on the prior hypothesis that negative samples (not being clustered) lie inside the convex discrimination polyhedron. This may be a limitation when it does not hold for a given data-set (left examples of Fig. \ref{fig:toy_examples_configuration_table}). Another hypothesis is that relevant psychiatric subtypes should not be based on the disease severity. This a priori implies that clusters should be along the classification boundary (upper examples of Fig. \ref{fig:toy_examples_configuration_table}). Even though it may help circumvent general variability issues, this strongly limits the applicability of the method to a specific variety of subgroups.

\subsubsection{Contributions}
Here, we propose a general framework for Unsupervised Clustering driven by Supervised Learning (UCSL) for relevant subtypes discovery.
The estimate of the latent subtypes is tied with the supervision task (regression or classification). 
Furthermore, we also propose to use an ensembling method in order to avoid trivial local minima or collapsed clusters.

We demonstrate the relevance of the UCSL framework on several data-sets.
The quality of the obtained results, the high versatility, and the computational efficiency of the proposed framework make it a good
choice for many subtype discovery applications in various domains.
Additionally, the proposed method needs very few parameters compared to other state-of-the-art (SOTA) techniques, making it more relevant for a large number of medical applications where the number of training samples is usually limited.
Our three main contributions are :
\begin{enumerate}
    \item A generic mathematical formulation for subtype discovery which is robust to samples inside and outside the classification polytope (see Fig.\ref{fig:toy_examples_configuration_table}).
    \item An Expectation-Maximization (EM) algorithm with an efficient dimensionality reduction technique during the E step for estimating latent subtypes more relevant to the supervised task. 
    \item A thoughtful evaluation of our subtype discovery method and a fair comparison with several other SOTA techniques on both synthetic and real data-sets. In particular, a neuroimaging data-set for psychiatric subtype discovery.
\end{enumerate}

\section{UCSL: an Unsupervised Clustering driven by Supervised Learning framework}
\subsection{Mathematical formulation}
Let $(X,Y)=\{(x_i,y_i) \}_{i=1}^n$ be a labeled data-set composed of $n$ samples. Here, we will restrict to regression, $y_i \in \mathcal{R}$, or binary classification, $y_i \in \{-1,+1 \}$. We assume that all samples, or only positive samples ($y_i=+1$), can be subdivided into latent subgroups for regression and binary classification respectively.
\newline The membership of each sample $i$ to latent clusters is modeled via a latent variable $c_i \in C=\{C_1,...,C_K\}$, where $K$ is the number of assumed subgroups. We look for a discriminative model that maximizes the joint conditional likelihood:
\begin{equation}
    \sum_{i=1}^n \log \sum_{c \in C} p(y_i, c_i  | x_i)
\end{equation}

Directly maximizing this equation is hard and it would not explicitly make the supervised task and the clustering depend on each other, namely we would like to optimize both $p(c_i|x_i,y_i)$ (the clustering task) and $p(y_i|x_i,c_i)$ (the upstream supervised task) and not only one of them. To this end, we introduce $Q$, a probability distribution over $C$, so that $\sum_{c_i \in C} Q(c_i)=1$.

\begin{equation}
    \sum_{i=1}^n \log \sum_{c \in C} p(y_i, c_i  | x_i) = \sum_{i=1}^n \log \left( \sum_{c \in C}  Q(c_i) \frac{p(y_i, c_i  | x_i)}{Q(c_i)} \right).
\end{equation}
\newline

\noindent By applying the Jensen inequality, we then obtain the following lower-bound:

\begin{equation}
    \sum_{i=1}^n \log \left( \sum_{c \in C}  Q(c_i) \frac{p(y_i, c_i  | x_i)}{Q(c_i)} \right) \geq \sum_{i=1}^n \sum_{c \in C} Q(c_i) \log \left( \frac{p(y_i, c_i  | x_i)}{Q(c_i)} \right),
    \label{eq:3}
\end{equation}

\noindent It can be shown that equality holds when:
\begin{equation}
    Q(c_i) = \frac{p(y_i, c_i | x_i)}{ \sum_{c \in C} p(y_i,c_i | x_i)} = \frac{p(y_i, c_i | x_i)}{p(y_i | x_i)} = p(c_i | y_i, x_i).
    \label{eq:4}
\end{equation}

\noindent The right term of Eq.~\ref{eq:3} can be re-written as:
\begin{equation}
     \sum_{i=1}^n  \sum_{c \in C} \bigg(Q(c_i) \log \Big(p(y_i | c_i, x_i)p(c_i | x_i) \Big) - Q(c_i) \log Q(c_i) \bigg).
\label{eq:5}
\end{equation}

We address the maximization of Eq.~\ref{eq:5} with an EM optimization scheme (algo.~\ref{alg:UCSL_algorithm}) that exploits linear models to drive the clustering until we obtain a stable solution. First, during the Expectation step, we tighten the lower bound in Eq.~\ref{eq:3} by estimating $Q$ as the latent clusters conditional probability distribution $p(c_i | y_i, x_i)$ as in Eq.~\ref{eq:4}. Then, we fix $Q$, and maximize the supervised conditional probability distribution $p(y_i | c_i, x_i)$ weighted by the conditional cluster distribution $p(c_i | x_i)$ as in Eq.~\ref{eq:5}. 

\subsection{Expectation step}
In this step, we want to estimate $Q$ as $p(c_i | y_i, x_i), \forall i \in \llbracket 1, n \rrbracket, \forall c \in C $ in order to tighten the lower bound in Eq.~\ref{eq:3}. 
We remind here that latent clusters $c$ are defined only for the positive samples ($y=+1$), when dealing with a binary classification, and for all samples in case of regression.
Let us focus here on the binary classification task. Depending on the problem one wants to solve, different solutions are possible. On the one hand, if ground truth labels for classification are \textit{not} available at inference time, $Q$ should be computed using the classification prediction. For example, one could use a clustering algorithm only on the samples predicted as positive. However, this would bring a new source of uncertainty and error in the subgroups discovery due to possible classification errors. On the other hand, if ground truth labels for classification are available at inference time, one would compute the clustering using only the samples associated to ground-truth positive labels $\tilde{y}_i = +1$, and use the classification directions to guide the clustering. Here, we will focus on the latter situation, since it's of interest for many medical applications.

\begin{figure}[!tbp]
\includegraphics[width=0.95\textwidth]{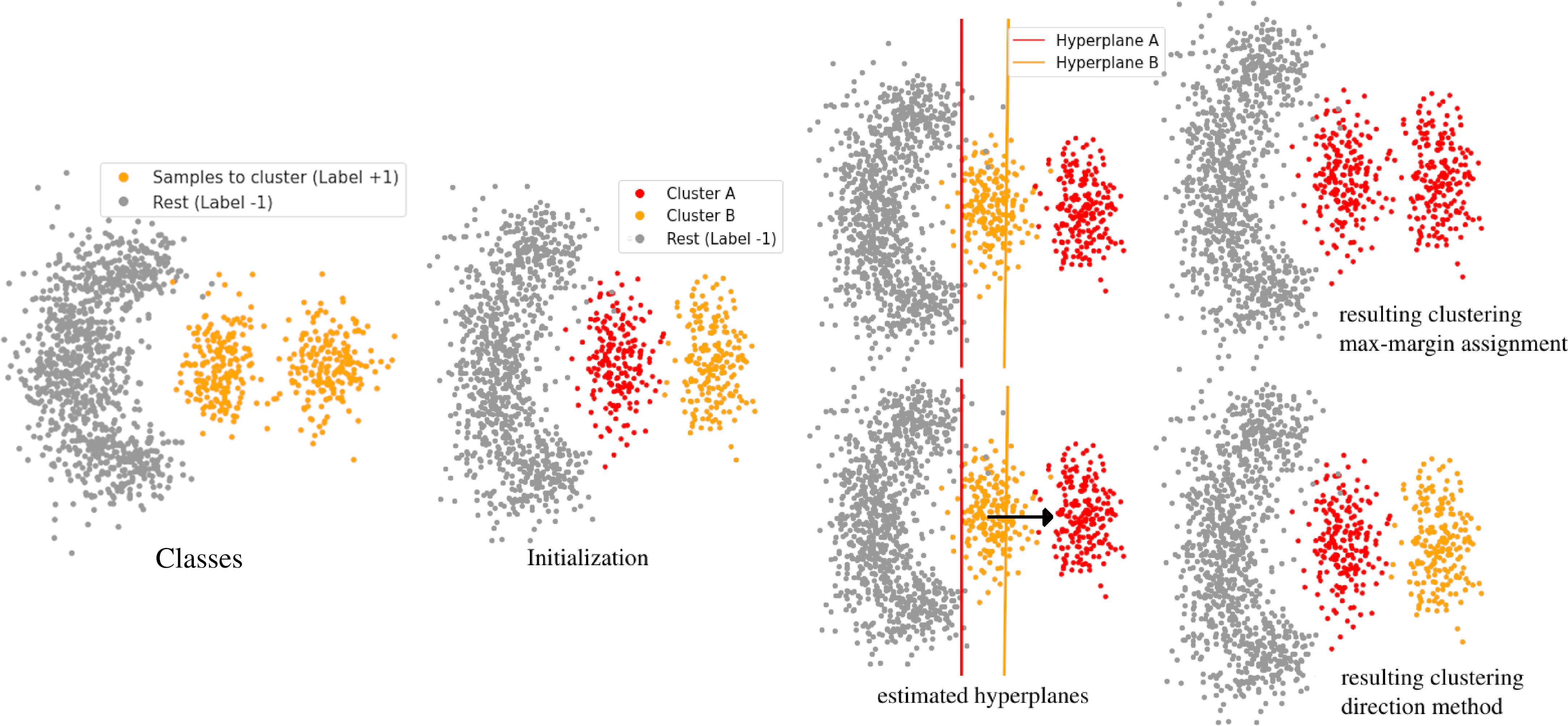}
\caption{Limit of maximum-margin based clustering starting from an optimal cluster initialization. When the separation of clusters to discover is co-linear to the supervised classification boundary, the maximum margin cluster assignment (as in \cite{HYDRA}) converges towards a degenerate solution (upper figures). Instead, with our direction method (lower figures), the Graam-Schmidt algorithm returns one direction where input points are projected to and perfectly clustered.}
\label{fig:HYDRA_clustering_limit}
\end{figure}

Now, different choices are again possible. In order to influence the resulting clustering with the label prediction estimation, HYDRA \cite{HYDRA} proposes to assign each positive sample to the hyperplane that best separates it from negative samples (i.e. the furthest one). This is a simple way to align resulting clustering with estimated classification while implicitly leveraging  classification boundary non-linearity.
Yet, we argue that this formulation does not work in the case where clusters are disposed parallel to the piece-wise boundary as described in Fig. \ref{fig:HYDRA_clustering_limit}. To overcome this limit, we propose to project input samples onto a supervision-relevant subspace before applying a general clustering algorithm.

\subsubsection{Dimension reduction method based on discriminative directions}
Our goal is a clustering that best aligns with the upstream-task. In other words, in a classification example, the discovery of subtypes should focus on the same features that best discriminate classes, and not on the ones characterizing the general variability. In regression, subgroups should be found by exploiting features that are relevant for the prediction task.
In order to do that, we rely on the linear models estimated from the maximization step. More specifically, we propose to first create a relevant orthonormal sub-space by applying the Graam-Schmidt algorithm onto all discriminant directions, namely the normal directions of estimated hyperplanes. Then, we project input features onto this new linear subspace to reduce the dimension and perform cluster analysis on a more suitable space. Clustering can be conducted with any algorithm such as Gaussian Mixture Models (GMM), K-Means (KM) or DBSCAN for example.


\begin{algorithm}
	\caption{Dimension reduction method based on discriminative directions}
    \label{alg:graam_schmidt}
    \hspace*{\algorithmicindent} \textbf{Input :} $X \in \mathbf{R}^{n \text{x} d} $, training data with $n$ samples and $d$ features. \\
    \hspace*{\algorithmicindent} \textbf{Output :} $X' \in \mathbf{R}^{n \text{x} K} $, training data projected onto relevant orthonormal subspace.
	\begin{algorithmic}[1]
	\State Given $K$ estimated hyperplanes, concatenate normal vectors in $D \in \mathbf{R}^{K \text{x} d}$.
	\State Ortho-normalize the direction basis $D$ with Graam-Schmidt obtaining $D^{\perp} \in \mathbf{R}^{K \text{x} d}$.
	\State Project training data onto the orthonormal subspace, $X' = X(D^{\perp})^T $ .
	\end{algorithmic} 
\end{algorithm}

\subsection{Maximization step}
After the expectation step, we fix $Q$ and then maximize the conditional likelihood. The lower bound in Eq.~\ref{eq:5} thus becomes:
\begin{equation}
    \label{eq:6}
    \sum_{i=1}^n \sum_{c \in C} Q(c_i) \log p(y_i | c_i, x_i) + \sum_{i=1}^n \sum_{c \in C} Q(c_i) \log p(c_i | x_i)
\end{equation}

\begin{figure}[!tbp]
\centering
\includegraphics[width=\textwidth]{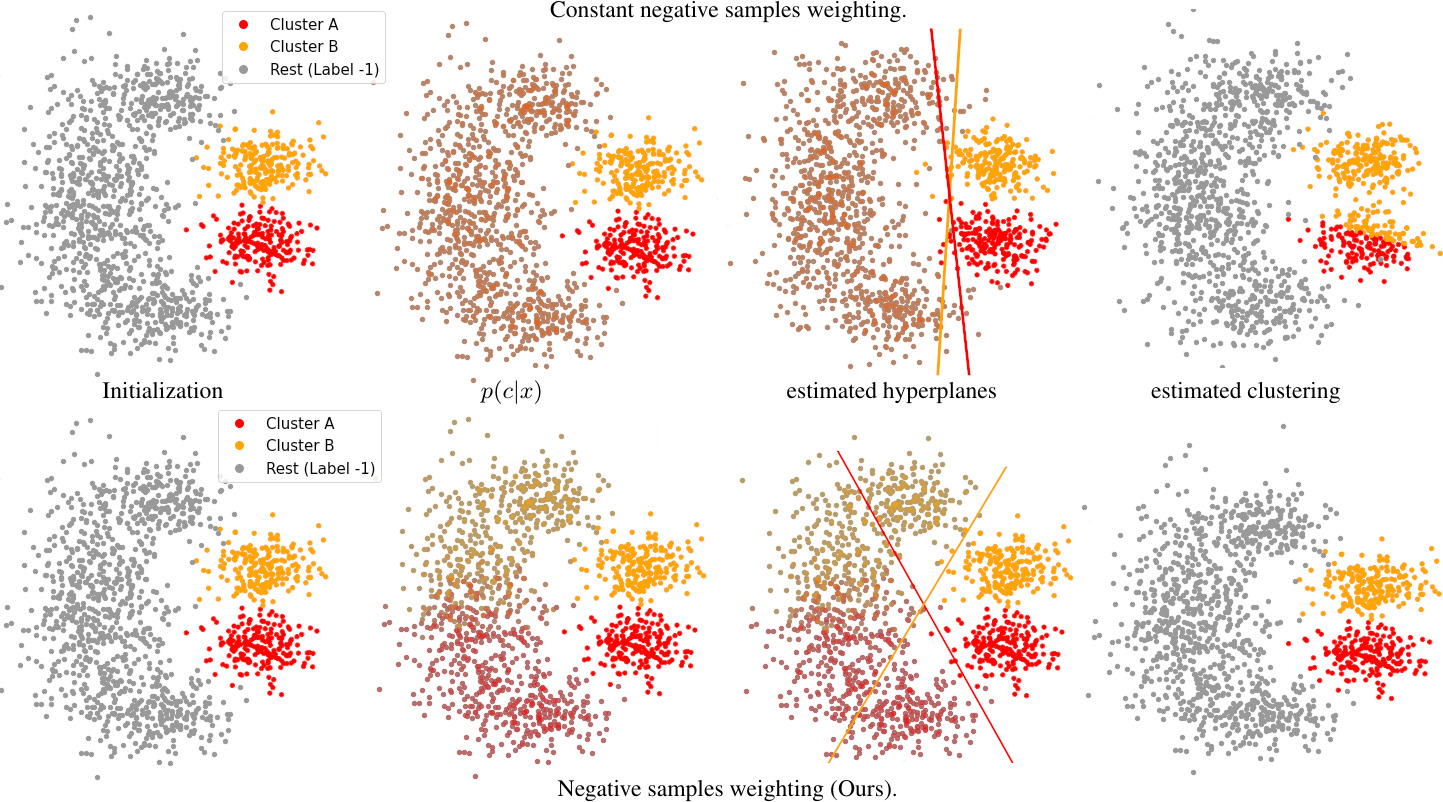}
\caption{
Starting with an optimal initialization of clusters to discover, constant negative samples weighting (top row) may lead to co-linear discriminative hyperplanes and thus errors in clustering. Conversely, our negative samples weighting enforces non-colinearity between discriminative hyperplanes resulting in higher quality clustering.}
\label{fig:neg_weighting_comparison}
\end{figure}

Here, we need to estimate $p(c_i | x_i)$. 
A possible solution, inspired by HYDRA \cite{HYDRA}, would be to use the previously estimated distribution $p(c_i | y_i, x_i)$ for the positive samples and a fixed weight for the negative samples, namely:
\begin{equation}
    p(c_i | x_i) = 
    \left\{
        \begin{array}{ll}
            p(c_i | x_i, y_i) & \text{ if } \Tilde{y_i}=+1 \\
            \frac{1}{K} & \text{ if } \Tilde{y_i}=-1
        \end{array}
    \right.
\end{equation}
However, as illustrated in Fig. \ref{fig:neg_weighting_comparison}, this approach does not work well  when negative samples lie outside of the convex classification polytope since discriminative directions (or hyperplanes) may become collinear. This collinearity hinders the retrieving of informative directions and consequently degrades the resulting clustering. 

\noindent To overcome such a shortcoming, we propose to approximate 
$p(c_i | x_i)$ using $p(c_i | x_i, y_i)$ for both negative and positive samples or, in other words, to extend the estimated clustering distribution to all samples, regardless their label $y$. In this way, samples from the negative class ($y_i = -1$), that are closer to a certain positive cluster, will have a higher weight during classification. As shown in Fig. \ref{fig:neg_weighting_comparison}, this results in classifications hyperplanes that correctly separate each cluster from the closer samples of the negative class, entailing  better clustering results.
From a practical point of view, since we estimate $Q(c_i)$ as $p(c_i| y_i, x_i)$, it means that $p(c_i | x_i)$ can be approximated by $Q(c_i)$. $Q(c_i)$ being fixed during the M step, only the left term in Eq.~\ref{eq:6} is maximized.


\subsection{Supervised predictions}
Once trained the proposed model, we compute the label $y_j$ for each test sample $x_j$ using the estimated  conditional distributions  $p(y_j | c_j, x_j)$ and $p(c_j | x_j)$ as:

\begin{equation}
 p(y_j | x_j) = \sum_{c_j \in C}  p(y_j,c_j | x_j) = \sum_{c_j \in C} p(y_j | x_j, c_j) p(c_j | x_j)    
\end{equation}

In this way, we obtain a non-linear estimator based on linear hyperplanes, one for each cluster.


\subsection{Application}
\subsubsection{Multiclass case}
In the case of classification, we handle the binary case in the same way as \cite{HYDRA} does. We consider one label as positive $\Tilde{y_i}=1$ and cluster it with respect to the other one $\Tilde{y_i}=-1$. In the multi-class case, we can cast it as several binary problems using the one-vs-rest strategy. 

\subsubsection{Ensembling : Spectral clustering} \label{ensembling}
The consensus step enables the merging of several different clustering propositions to obtain an aggregate clustering. After having run the EM iterations $N$ times, the consensus clustering is computed by grouping together samples that were assigned to the same cluster across different runs. In practice, we compute a co-occurrence matrix between all samples. And then we use co-occurence values as a similarity measure to perform spectral clustering. Hence, for example, given two samples $i$ and $j$ and $10$ different runs, if samples $i$ and $j$ ended up $4$ times in the same cluster, the similarity measure between those 2 samples will be $\frac{4}{10}$. Given an affinity matrix between all samples, we can then use the spectral clustering algorithm to obtain a consensus clustering.

\subsection{Pseudo-code}
The pseudo-code of the proposed method UCSL (Alg. \ref{alg:UCSL_algorithm}) can be subdivided into several distinct steps:


\begin{algorithm}
    \hspace*{\algorithmicindent} \textbf{Input :} $X \in \mathbf{R}^{n \text{x} d} $, $y \in \{-1, 1\}^n$, $K$ number of clusters. \\
    \hspace*{\algorithmicindent} \textbf{Output :} $p(c | x, y)=Q(c)$, $p(y | x, c)$ (linear sub-classifiers).
	\begin{algorithmic}[1]
	\For {ensemble in n\textunderscore ensembles}
	    \State {\bf Initialization}: Estimate $Q^{(0)}$ for all samples ($y=\pm 1$) with a clustering algorithm (e.g. GMM) trained with positive samples only ($y=+1$).
	    \While {not converged}
   		    \State {\bf M step} (supervised step) :
   	    \State \hspace*{\algorithmicindent} Freeze $Q^{(t)}$ 
   	        \State \hspace*{\algorithmicindent} {\bf for} {k in $[1, K]$} :
   		    \State \hspace*{\algorithmicindent} \hspace*{\algorithmicindent} Fit linear sub-classifier $k$ weighted by $Q^{(t)}[:,k]$ (Eq.~\ref{eq:6}).
   		    \State \hspace*{\algorithmicindent} {\bf end for}
    		\State {\bf E step} (unsupervised step) :
   		    \State \hspace*{\algorithmicindent} Use Alg.\ref{alg:graam_schmidt} to obtain $X' \in \mathbf{R}^{n \text{x} K}$ from sub-classifiers normal vectors $D \in \mathbf{R}^{K \text{x} d}$.
   		    \State \hspace*{\algorithmicindent} Estimate $Q^{(t+1)} = p(c | x, y)$ (Eq.~\ref{eq:4}) for all samples with a clustering algorithm trained on $X'$ with positive samples only.
    	\EndWhile
    \EndFor
    \State {\bf Ensembling}: Compute average clustering with the ensembling method (Sec.~\ref{ensembling}).
        \State {\bf Last EM} : Perform EM iterations from ensembled latent clusters until convergence.
	\end{algorithmic} 
\caption{UCSL general framework pseudo-code}
\label{alg:UCSL_algorithm}
\end{algorithm} 

\begin{enumerate}
    \item {\bf Initialization:} First, we have to initialize the clustering. There are several possibilities here, we can make use of traditional ML methods such as KM or GMM. For most of our experiments we used GMM.
    \item {\bf Maximization:} The Maximization step consists in training several linear models to solve the supervised upstream problem.
    It can be either a classification or a regression. We opted for well-known ML linear methods such as logistic regression or max-margin linear classification method as in \cite{HYDRA}.
    \item {\bf Expectation:} The Expectation step makes use of the supervised learning estimates to produce a relevant clustering. In our case, we exploit the directions exhibited by the linear supervised models. We project samples onto a subspace spanned by those directions to perform the unsupervised clustering with positive samples.
    \item {\bf Convergence:} In order to check the convergence, we compute successive clustering Adjusted Rand Score (ARI), the closer this metric is to 1, the more similar both clustering assignments are.
    \item {\bf Ensembling:} Initialization and EM iterations are performed until convergence $N$ times and an average clustering is computed with a Spectral Clustering algorithm \cite{HYDRA}, \cite{planey_coincide_2016} that proposes the best consensus. This part enables us to have more robust and stable solutions avoiding trivial or degenerate clusters.
\end{enumerate}

\section{Results}

We validated our framework on four synthetic data-sets and two  experimental ones both qualitatively and quantitatively. 

\subsubsection{Implementation details}
The stopping criteria in Alg. \ref{alg:UCSL_algorithm} is defined using the ARI index between two successive clusterings (at iteration $t$ and iteration $t+1$). The algorithm stops when it reaches the value of $0.85$.
In the MNIST experiment, convolutional generator and encoder networks have a similar structure to the generator and discriminator in DCGAN \cite{radford_unsupervised_2016}. We trained it during 20 epochs, with a batch size of 128, a learning rate of 0.001 and with no data augmentation and a SmoothL1 loss. More information can be found in the Supplementary material. Standard deviations are obtained by running 5 times the experiments with different initializations (synthetic and MNIST examples) or using a 5-fold cross-validation (psychiatric dataset experiment).
MNIST and synthetic examples were run on Google Colaboratory Pro, whose hardware equipments are PNY Tesla P100 with 28Gb of RAM.

\subsubsection{Synthetic dataset}
\begin{figure}[!tbp]
\centering
\includegraphics[width=\textwidth]{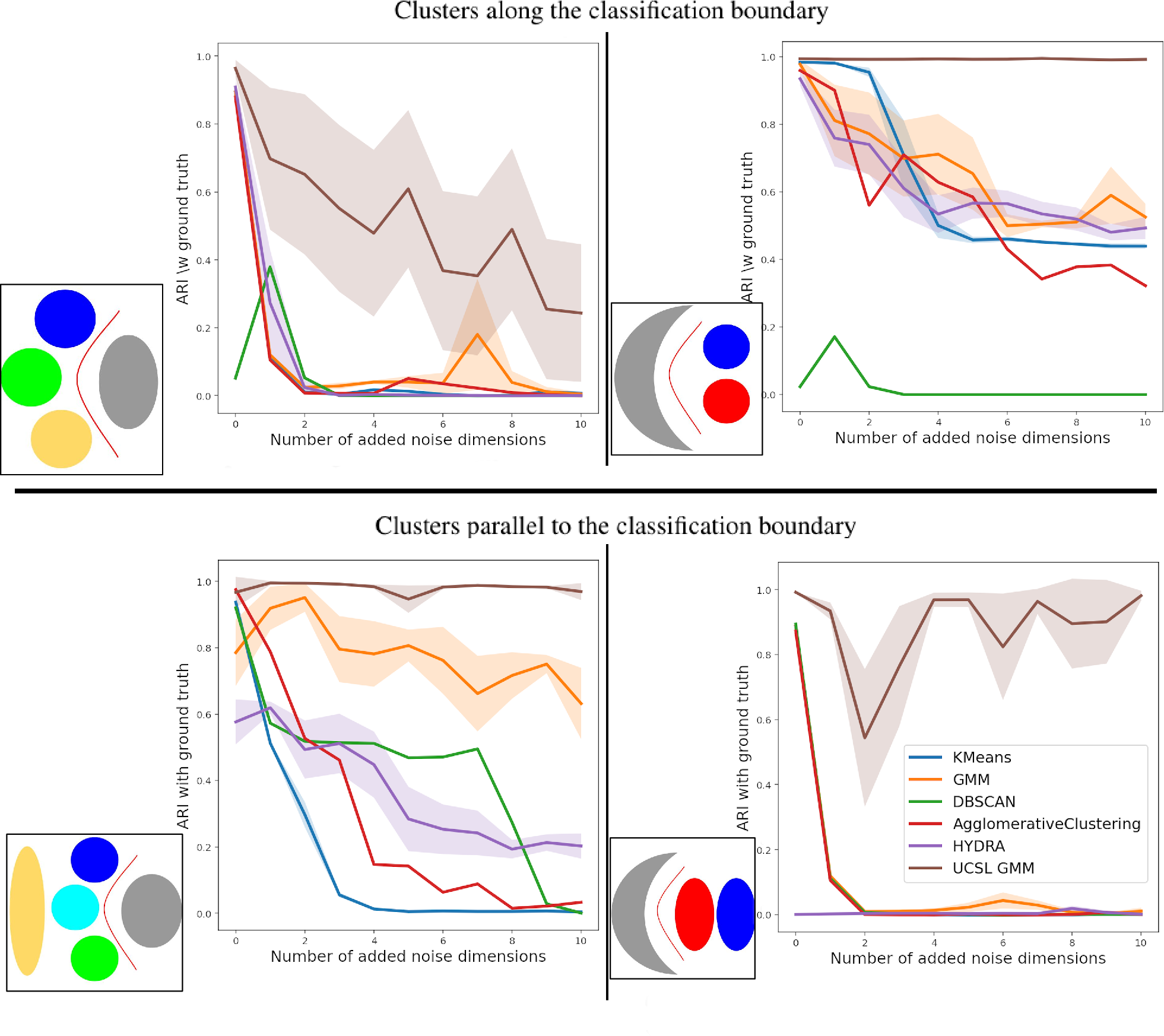}
\centering
\caption{
Comparison of performances of different algorithms on the four configurations presented in Fig.\ref{fig:toy_examples_configuration_table}. Noisy features are added to the original 2D data. For each example, all algorithms are run 10 times with different initialization.} 
\label{fig:toy_examples_performance_table}
\end{figure}

First, we generated a set of synthetic examples that sum up the different configurations on which we wish our method to be robust: subtypes along the supervised boundary or parallel to it. We designed configurations with various number of clusters, outside or inside the convex classification polytope. 
UCSL was run with Logistic Regression and GMM.
In order to make our problem more difficult we decided to add noisy unnecessary features to the original 2-D toy examples. For each example and algorithm, we performed 10 runs with a different initialization each time (GMM with only one initialization) and we did not perform the ensembling step for fair comparison with the other methods. We compared with other traditional ML methods such as KM GMM, DBSCAN and Agglomerative Clustering. Results are displayed in Fig.~\ref{fig:toy_examples_performance_table}. For readability, we divided the standard deviation hull by 2. Compared with the other methods, UCSL appears to be robust to unnecessary noisy features. Furthermore, it performs well in all configurations we addressed. 

\subsubsection{MNIST dataset}
To further demonstrate what an intra-class clustering could be used for, let us make an example from MNIST. We decided to analyse the digit 7 looking for subtypes. To perform this experiment, we trained on 20 000 MNIST digits and considered the digit 7 as positive class. We use a one-vs-rest strategy for classification where input samples are the flattened images. 

Visually, digit 7 examples have two different subtypes: with or without the middle-cross bar. In order to quantitatively evaluate our method, we labeled 400 test images in two classes, 7 with a middle-cross bar, and those with none. We ran UCSL with GMM as a clustering method, logistic regression as classification method and compared with clustering methods coupled with deep learning models or dimension reduction algorithms. We use the metrics V-Measure, Adjusted Rand Index (ARI) and balanced accuracy (B-ACC), since we know the expected clustering result. 

\begin{table}[h!]
\centering
\begin{tabular}{ |c|c|c|c||c|c|c|c| } 
\hline
Methods & Latent Size & Nb params & Avg Exec Time & V-measure & ARI & B-ACC \\
\hline
AE + GMM & 32 & 3M & 21m40s & 0.323$\pm$0.013 & 0.217$\pm$0.025 & 0.823$\pm$0.009 \\
UCSL (our) & 2 & 406 & 12m31s & 0.239$\pm$0.001 & 0.330$\pm$0.001 & \textbf{0.808$\pm$0.001} \\
PT VGG11 + KM & 1000 & 143M & 32m44s & 0.036$\pm$0.001 & 0.087$\pm$0.001 & 0.616$\pm$0.001 \\
AE + GMM & 2 & 3M & 13m34s & 0.031$\pm$0.015 & 0.033$\pm$0.021 & 0.607$\pm$0.025 \\
t-sne* \cite{maaten_visualizing_2008} + KM & 2 & 4 & 2m04s & 0.029$\pm$0.020 & 0.049$\pm$0.056 & 0.568$\pm$0.033 \\ 
t-sne* \cite{maaten_visualizing_2008} + GMM & 2 & 14 & 2m04s & 0.023$\pm$0.021 & 0.020$\pm$0.048 & 0.566$\pm$0.033 \\
umap* \cite{mcinnes_umap_2020} + KM & 2 & 4* & 24s & 0.050$\pm$0.015 & 0.078$\pm$0.015 & 0.555$\pm$0.022 \\
umap* \cite{mcinnes_umap_2020} + GMM & 2 & 14* & 24s & 0.025$\pm$0.006 & 0.080$\pm$0.010 & 0.547$\pm$0.005 \\
SHAP \cite{lundberg_consistent_2017}* + KM & 196 & 392* & 1h02 & 0.012$\pm$0.007 & -0.014$\pm$0.035 & 0.540$\pm$0.016 \\
KM & 196 & 392 & 0.32ms & 0.006$\pm$0.000 & 0.010$\pm$0.000 & 0.552$\pm$0.000 \\ 
HYDRA & 196 & 394 & 9m45s & 0.005+/-0.006 & 0.024$\pm$0.031 & 0.520$\pm$0.018 \\
GMM & 196 & 77K & 0.32ms & 0.0002$\pm$0.000 & -0.001$\pm$0.000 & 0.510$\pm$0.000 \\
\hline
\end{tabular}
\newline
\caption{MNIST dataset, comparison of performances of different algorithms for the discovery of digit 7 subgroups. AE : convolutional AutoEncoder; PT VGG11: VGG11 model pre-trained on imagenet; GMM: Gaussian Mixture Model; KM: K-Means. Latent size: dimension of space where clustering is computed.
*  : to limit confusion, we assign no parameters for t-sne, umap and SHAP.
We use default values (15,30,100) for perplexity, neighbours and n estimators in t-sne, umap and SHAP respectively. }
\label{table:7_digit_subtype_discovery_table}
\end{table}

\begin{figure}[!tbp]
  \centering
  \begin{minipage}[b]{0.49\textwidth}
    \subcaption{UCSL (Ours)}
    \includegraphics[width=\textwidth]{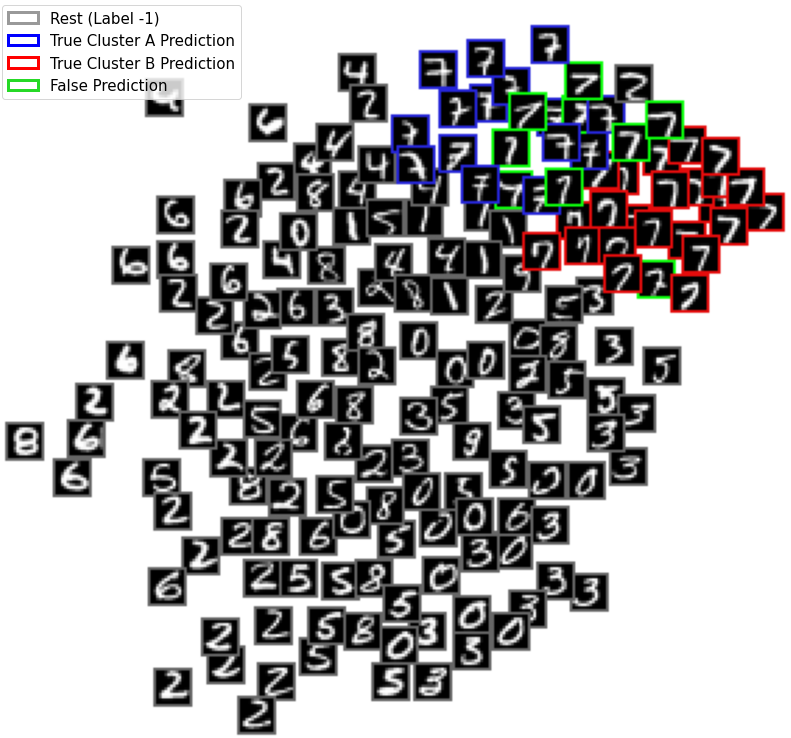}
  \end{minipage}
  \hfill
  \begin{minipage}[b]{0.49\textwidth}
    \subcaption{t-SNE \cite{maaten_visualizing_2008} + KMeans}
    \includegraphics[width=\textwidth]{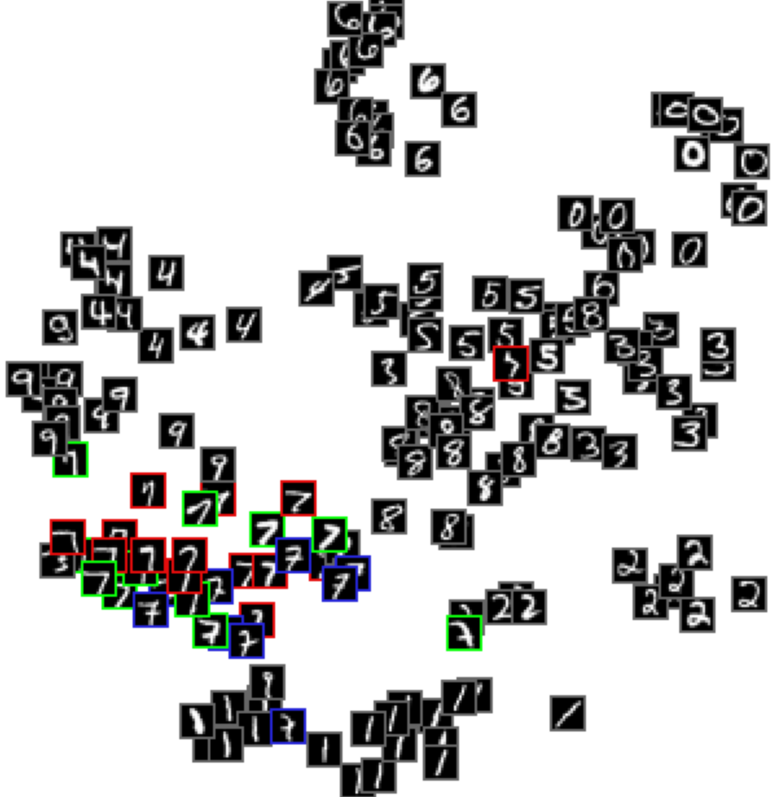}
  \end{minipage}
\caption{Comparison of latent space visualization in the context of MNIST digit ``7" subtype discovery. Differently from t-SNE, our method does not focus on the general digits variability but only on the variability of the ``7". For this reason, subtypes of ``7" are better highlighted with our method. }
\label{table:MNIST_7_digit_clustering}
\end{figure}

As it is possible to notice from Table \ref{table:7_digit_subtype_discovery_table}, UCSL outperforms other clustering and subtypes ML methods. We also compared our algorithm with DL methods, a pre-trained convolutional network and a simple convolutional encoder-decoder. Only the convolutional autoencoder network along with a GMM on its latent space of dimension 32 slightly outperforms UCSL. However, it uses a definitely higher number of parameters (7500 times more!) and takes twice the time for training. Our model is thus more relevant to smaller data-sets, which are common in medical applications. Please note that UCSL could also be adapted in order to use convolutional auto-encoders or contrastive methods such as in \cite{li_prototypical_2021} and \cite{caron_deep_2018}, when dealing with large data-sets. This is left as future work. 

\subsubsection{Psychiatric dataset}
The ultimate goal of the development of subtype discovery methods is to identify homogeneous subgroups of patients that are associated with different disease mechanisms and lead to patient-specific treatments. With brain imaging data, the variability specific to the disorder is mixed up or hidden to non-specific variability. Classical clustering algorithms produce clusters that correspond to subgroups of the general population: old participants with brain atrophy versus young participants without atrophy, for instance.

To validate the proposed method we pooled neuroimaging data from patients with two psychiatric disorders, (Bipolar Disorder (BD) and Schizophrenia (SZ)), with data from healthy controls (HC). The supervised upstream task aims at classifying HC from patients (of both disorders) using neuroimaging features related to the local volumes of brain grey matter measured in 142 regions of interest (identified using cat12 software). Here, we used a linear SVM for classification. The clustering task is expected to retrieve the known clinical disorder (BD or SZ). Training set was composed of 686 HC and 275 SZ, 307 BP patients.

We measured the correspondence (Tab.~\ref{table:scz_vs_bip_ML_results}) between the clusters found by the unsupervised methods with the true clinical labels on an independant TEST set (199 HC, 190 SZ, 116 BP) coming from a different acquisition site. As before, we used the metrics V-Measure, Adjusted Rand Index (ARI) and balanced accuracy (B-ACC). Please note that the classification of SZ vs BD is a very difficult problem due to the continuum between BP and SZ.
Therefore, performances should be compared with the best expected result provided by a purely supervised model (here a SVM) that produces only 61\% of accuracy (last row of Tab.~\ref{table:scz_vs_bip_ML_results}).

\begin{table}[ht!]
\centering
\setlength\tabcolsep{0.035\textwidth}
\begin{tabular}{ |c||c|c|c| } 
\hline
Algorithm & V-measure & ARI & B-ACC \\
\hline
GMM & 0.002$\pm$0.001 & 0.003$\pm$0.008 & 0.491$\pm$0.024 \\ 
KMeans & 0.008$\pm$0.001 & -0.01$\pm$0.001 & 0.499$\pm$0.029 \\ 
umap* \cite{mcinnes_umap_2020} + GMM & 0.001$\pm$0.002 & 0.000$\pm$0.007 & 0.497$\pm$0.013 \\
umap* \cite{mcinnes_umap_2020} + KM & 0.000$\pm$0.002 & 0.001$\pm$0.005 & 0.502$\pm$0.006 \\
t-sne* \cite{maaten_visualizing_2008} + GMM & 0.002$\pm$0.0024 & -0.00$\pm$0.005 & 0.498$\pm$0.028 \\
t-sne* \cite{maaten_visualizing_2008} + KM & 0.004$\pm$0.004 & 0.003$\pm$0.005 & 0.505$\pm$0.041 \\
HYDRA \cite{HYDRA} & 0.018$\pm$0.009 & -0.01$\pm$0.004 & 0.556$\pm$0.019 \\
SHAP \cite{schulz_inferring_2020} + GMM & 0.004$\pm$0.005 & 0.000$\pm$0.006 & 0.527$\pm$0.027 \\
SHAP \cite{schulz_inferring_2020} + KMeans & 0.016$\pm$0.005 & 0.017$\pm$0.012 & 0.575$\pm$0.011 \\
UCSL + GMM & \textbf{0.024$\pm$0.006} & \textbf{0.042$\pm$0.016} & \textbf{0.587$\pm$0.009} \\
UCSL + KMeans & \textbf{0.030$\pm$0.012} & 0.004$\pm$0.006 & \textbf{0.594$\pm$0.015} \\ 
\hline
\hline
\textcolor{gray}{\textit{Supervised SVM}} & \textcolor{gray}{\textit{0.041$\pm$0.007}} & \textcolor{gray}{\textit{0.030$\pm$0.008}} & \textcolor{gray}{\textit{0.617$\pm$0.010}} \\
\hline
\end{tabular}
\newline
\caption{Results of the different algorithms on the subtype discovery task BP / SZ. The last row provides the best expected result obtained with a supervised SVM.}
\label{table:scz_vs_bip_ML_results}
\end{table}

As expected, mere clustering methods (KMeans, GMM) provide clustering at the chance level. Detailed inspection showed that they retreived old patients with brain atrophy vs younger patients without atrophy. Only clustering driven by supervised upstream task (HYDRA, SHAP+KMeans and all UCSL) can disentangle the variability related to the disorders to provide results that are significantly better than chance (59\% of B-ACC). Models based on USCL significantly outperformed all other models approaching the best expected result that would provide a purely supervised model.  

\section{Conclusion}
We proposed in this article a Machine Learning (ML) Subtype Discovery (SD) method that aims at finding relevant homogeneous subgroups with significant statistical differences in a given class or cohort. To address this problem, we introduce a general Subtype Discovery (SD) Expectation-Maximization (EM) ensembled framework. 
We call it UCSL : Unsupervised Clustering driven by Supervised Learning. Within the proposed framework, we also propose a dimension reduction method based on discriminative directions to project the input data onto an upstream-task relevant linear subspace.
UCSL is adaptable to both classification and regression tasks and can be used with any clustering method. Finally, we validated our method on synthetic toy examples, MNIST and a neuro-psychiatric 
data-set on which we outperformed previous state-of-the-art methods by about +1.9 points in terms of balanced accuracy.

%
%
%
\bibliographystyle{splncs04}
\bibliography{UCSL}

\end{document}